# DREaM: Drug-Drug Relation Extraction via Transfer Learning Method

Ali Fata, Hossein Rahmani*, Parinaz Soltanzadeh, Amirhossein Derakhshan, Behrouz Minaei Bidgoli

School of Computer Engineering

Iran University of Science and Technology, Tehran, Iran

ali_fata@comp.iust.ac.ir, h_rahmani@iust.ac.ir, p_soltanzadeh77@comp.iust.ac.ir, am_derakhshan@comp.iust.ac.ir, b_minaei@iust.ac.ir

*Abstract*— **Relation extraction between drugs plays a crucial role in identifying drug–drug interactions and predicting side effects. The advancement of machine learning methods in relation extraction, along with the development of large medical text databases, has enabled the low-cost extraction of such relations compared to other approaches that typically require expert knowledge. However, to the best of our knowledge, there are limited datasets specifically designed for drug–drug relation extraction currently available. Therefore, employing transfer learning becomes necessary to apply machine learning methods in this domain. In this study, we propose DREAM, a method that first employs a trained relation extraction model to discover relations between entities and then applies this model to a corpus of medical texts to construct an ontology of drug relationships. The extracted relations are subsequently validated using a large language model. Quantitative results indicate that the LLM agreed with 71 % of the relations extracted from a subset of PubMed abstracts. Furthermore, our qualitative analysis indicates that this approach can uncover ambiguities in the medical domain, highlighting the challenges inherent in relation extraction in this field.**

## I. INTRODUCTION

With the growing volume of textual data across various domains, processing this data and extracting valuable information from it has become increasingly important. In the medical field, large-scale textual databases such as PubMed, Embase, and BioMed Central have been developed [1]. These resources facilitate tasks such as drug-drug interaction [2-4] and drug-disease interaction [5] extraction, adverse drug reaction prediction [6-8], and the construction of medical knowledge graphs [9-13]. One of the practical tasks in this domain is the extraction of relationships between drugs, which can be used for the development of ontologies based on them. Ontologies refer to a structured set of concepts and the relationships among them, defined within a specific domain [14]. This structure enables the interpretation of information within a semantic and conceptual framework and allows for the hierarchical modeling of relationships among entities.

Relation extraction, as one of the key tasks in natural language processing (NLP), plays a crucial role in the construction of ontologies and knowledge graphs. The primary objective of this task is to identify the existence and type of relationship between named entities within a text [15]. Named entities are textual spans that refer to real-world objects, and various methods have been proposed for their extraction [16-20]. These entities may include names of people, locations, dates, organizations, products, and more. In the biomedical domain, focusing on domain-specific entities such as drugs, diseases, and side effects is often more beneficial than addressing all types of named entities. The relations between entities can vary widely depending on the context, ranging from general types such as content-container, member-collection, entity-destination, cause-effect, and message-topic, to more domain-specific ones. In the medical domain, relationships like component-whole and cause-effect are particularly important. Depending on the application, it is common to narrow down the scope of relation extraction to a relevant subset of relation types.

Various datasets have been developed for the task of relation extraction [21-24], enabling the training and evaluation of machine learning and artificial intelligence models in this area. However, many of these datasets are not domain-specific, and applying models trained on them to a particular domain often requires transfer learning. Transfer learning is a machine learning technique that involves leveraging a model trained in one domain to perform tasks in a different, previously unseen domain [25]. This approach is particularly useful in fields such as medicine, where collecting annotated training data is either challenging or costly. In this study, we employ the concept of transfer learning to extract relations from medical texts, with the ultimate goal of extracting a set of drug-drug relationships that can be represented as an ontology.

In the following sections, we begin by reviewing the research that has been conducted in this field. Then, in the proposed method section, the approach proposed in this study is presented. Next, in the experimental results section, we evaluate the effectiveness of the proposed method. Finally, in the conclusions and future work section, we summarize the study and outline potential directions for further development.

## II. RELATED WORK

In this section, we review the work conducted in the field of relation extraction and one of its main applications, ontology construction. The existing studies in this area can be categorized into three main groups: statistical methods, machine learning methods, and deep learning methods. Figure 1 illustrates the categorization of the existing approaches in this field.

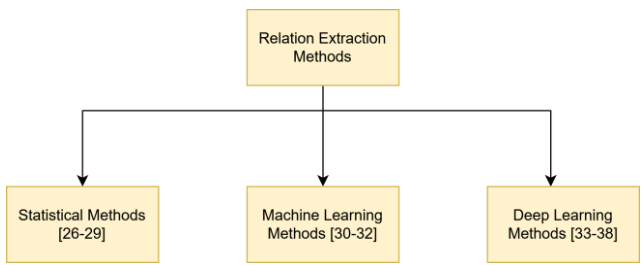

Figure 1. Classification of the existing studies in relation extraction task.

Statistical approaches uncover relationships within texts by first specifying pattern templates and then searching the corpus for segments that satisfy those templates. When a segment fits a template, the associated relation is recorded. Crafting these rule-based systems demands substantial subject-matter expertise and detailed linguistic insight to devise the extraction rules. They work especially well in narrowly defined domains where documents follow consistent formats and the relationships of interest are already known. Moving to a different domain, however, requires building an entirely new inventory of target relations and corresponding patterns. Comprehensive treatments of statistical techniques for relation extraction can be found in [26, 27], while statistical frameworks for building ontologies are surveyed in [28, 29].

Relation-extraction techniques grounded in machine learning fall into supervised, unsupervised, semi-supervised, and distant-supervised camps, all of which weave standard clustering or classification algorithms—such as K-means, logistic regression, or support-vector machines—into different steps of their workflows. In a distant-supervision model by Ranjbar et al [30] named entities are first detected, a unique "fingerprint" is generated for every candidate pair, and those fingerprints are matched against a knowledge base; query results and corpus statistics then gauge the likelihood of each pair and assign a relation label when warranted. Clustering-based variants [31, 32] adopt their own twists: one K-means approach ranks context terms by frequency and entropy to spotlight features before clustering, while Sekine's hierarchical scheme identifies keywords, groups similar ones, and ties those clusters to entity pairs. Graph-oriented methods instead exploit semantic similarity by constructing entity- or hypernym-level graphs, first discovering broad semantic classes that subsume many relations and then clustering together relations that appear semantically alike.

A typical neural-network pipeline for relation extraction has four stages: first, a lookup layer converts each token into a dense vector; next, a convolutional layer scans for informative n-gram patterns; third, pooling highlights the strongest signals; finally, a fully connected layer with soft-max assigns the relation label. Building on this template, Zhang et al [33] coupled a CNN feature extractor with a bidirectional LSTM that supplies wider sentence context, achieving entity and relation classification without handcrafted cues. Nayak et al [34] adapted encoder–decoder architectures for joint extraction, first letting the decoder spell out tuples token-by-token through a custom representation and later switching to a pointer-network variant that outputs an entire tuple in one step. Liu et al [35] boosted generalization by adding an auxiliary binary task that teaches the model to distinguish the synthetic "no relation" class from all genuine relations. Sui et al [36] recast joint extraction as predicting an unordered set and designed transformer decoders that emit every triple in parallel rather than sequentially. Geng et al [37] fused structural and sequential signals by passing dependency-tree representations through bidirectional Tree-LSTMs and token sequences through attentional LSTMs, then concatenating the results to improve accuracy. Zengeya et al [38] systematically review deep-learning models—RNNs, CNNs, LSTMs, and GRUs—for ontology construction, critiquing their strengths and weaknesses in term, entity, and relation extraction. They also assess advanced transformers such as GPT-3, GPT-4, and BERT, noting distinct advantages and limitations. The study follows a rigorous systematic-review protocol and outlines open research directions for deep learning in ontology building.

Rule based methods require expertise and knowledge in each field and their generalizability is usually low. Machine learning methods use statistical features to detect relationships, which eliminates the semantic dimension of words. However, these methods do not require high computational resources. Deep learning methods detect relationships by taking into account the semantic dimension of words. Although these methods have high accuracy, they generally require a large amount of training data and computational resources

## III. DREAM: PROPOSED METHOD

In this section, we present the method proposed in this study[1]. First, the relation extraction model employed in our approach is introduced. Then, we explain how this model is used to identify relationships between drugs. The architecture of the proposed method is illustrated in Figure 2.

### A. Relation extraction model

The relation extraction model employed in this study is the ACORD model, introduced in [39]. ACORD is a relation extraction approach that utilizes the ANOVA criterion to select the most important terms for each feature. It then uses an ensemble learning framework—consisting of Logistic Regression, Random Forest, and Gradient Boost models—to classify the relation based on the selected key terms. Compared to models based on neural networks, ACORD requires significantly fewer computational resources, enabling its broad and cost-effective application.

The ANOVA measure, which forms a key component of the ACORD model for feature selection, is used to identify the most discriminative terms for each relation type by analyzing the means and variances of the terms. The calculation of this measure between two classes is shown in Equation 1.

$$ANOVA = \frac{\sigma^2(\mu_1, \mu_2)}{\mu(\sigma_1^2, \sigma_2^2)} \tag{1}$$

In (1), $\mu$ refers to the mean of each class, and $\sigma$ represents the standard division of each class. Essentially, the larger the distance between the means of the two classes and the smaller

---

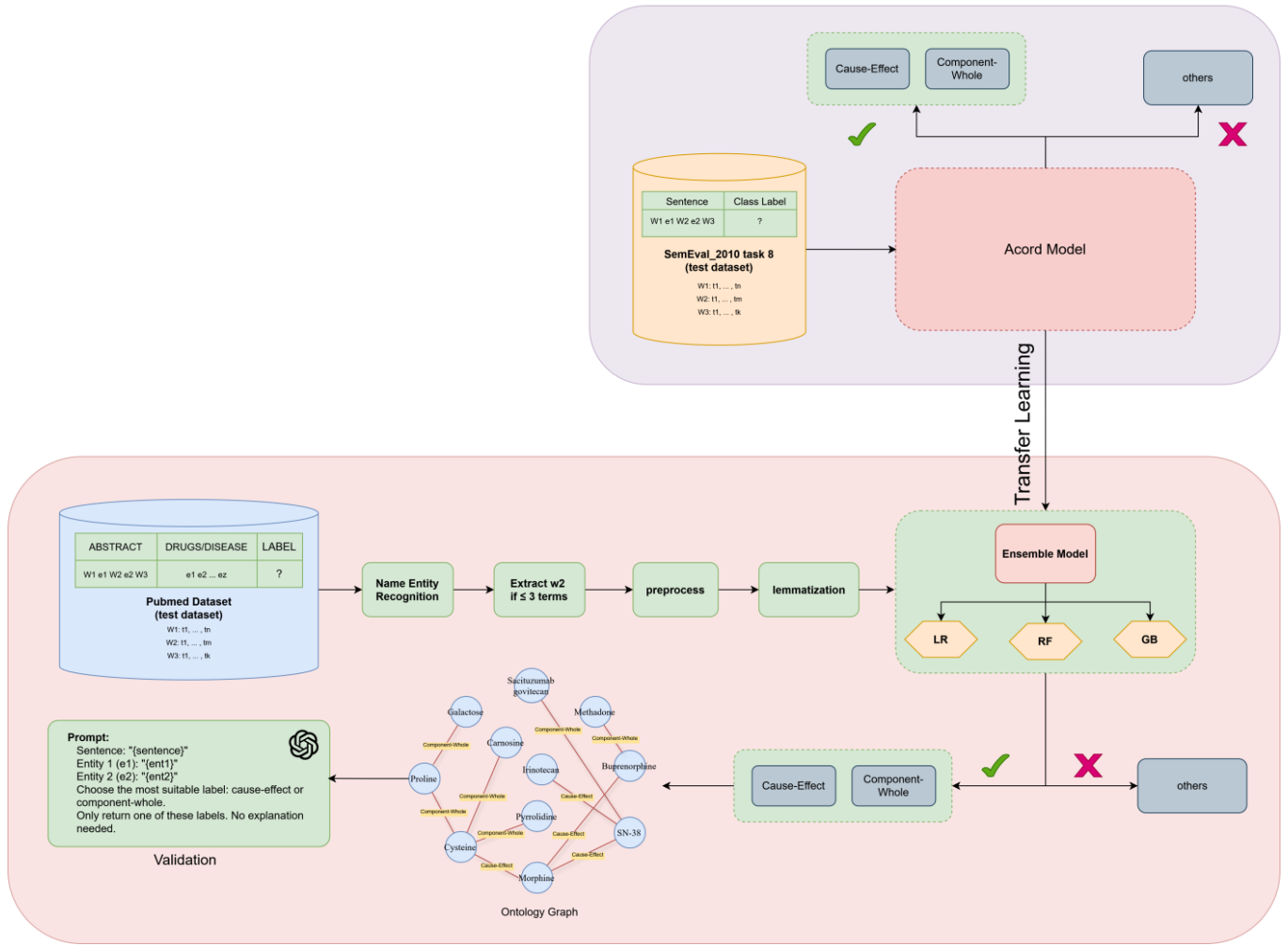

Figure 2. Architecture of the proposed method. (1) The ACORD model is trained to identify two types of relations: Cause-effect and Component-whole. (2) A collection of medical texts is gathered, and documents containing drug names are extracted using a predefined list. (3) The ACORD model is applied to the text between drug entities to identify a set of drug–drug relations. These extracted relations are then evaluated using an LLM.

the variance within the classes, the greater the ANOVA value becomes. A higher ANOVA value indicates better class separability.

### B. Drug-Drug relation extraction using transfer learning

To identify relationships between drugs, we targeted medical texts available in the PubMed database. PubMed is a free and reputable resource in the fields of life sciences and medicine, developed by the U.S. National Library of Medicine. Since the abstract of a paper typically contains its most important findings and information, we considered only the abstracts of a subset of articles published from 2022 onward. Accordingly, our goal is to detect drug-related relations within these abstracts.

In the next step, using the DrugBank dataset—which contains information about drugs and their functions—we selected abstracts that mentioned at least two drug names. For each such abstract, the text between the two drug names was provided to the ACORD model in order to extract the type of relation between them. If the identified relation was either *cause-effect* or *component-whole*, It was added to the set of relations that can be displayed as an ontology.

The decision to focus exclusively on *cause-effect* and *component-whole* relations is motivated by their critical role in ensuring the safety and efficacy of medical treatments. *Cause-effect* relations help identify drug-drug interactions, including adverse or synergistic effects, thereby preventing potentially dangerous side effects. On the other hand, *component-whole* relations are essential for understanding drug compositions, identifying active ingredients, and recognizing structural similarities among different drugs, which can support the design of new pharmaceuticals and the classification of similar drugs. The joint analysis of these two relation types provides a foundational basis for developing intelligent drug recommendation systems, uncovering hidden interactions, and optimizing combination therapies.

Finally, the resulting output is a graph $G = (V, E)$ in which $V$ denotes the set of drugs, and $E$, representing the relations between drugs, is defined by Equation 2.

$$E = \{(v_i, R, v_j) \mid v_i \in V \land v_j \in V \land R \in \\ \{cause-effect, component-whole\}\} \tag{2}$$

In Equation 2, $v_i$ and $v_j$ represent two nodes in the graph, corresponding to two drugs, and $R$ denotes the relation between them, which is either cause-effect or component-whole.

### C. Extracted Relations validation method

One of the main challenges in systems that aim to extract structured knowledge from text is evaluating the quality and correctness of the resulting representations. Various evaluation strategies have been introduced for this purpose. A common approach is gold standard comparison, where the extracted structure is evaluated against a predefined and trusted reference. While this method is straightforward and often reliable, it depends heavily on the availability of a high-quality gold standard. Another widely used strategy is corpus-based evaluation, where the extracted knowledge is compared against information derived from a relevant text corpus. This may involve checking for overlap in entities or using vector-based representations to assess similarity. Task-based evaluation is another method, which assesses the utility of the extracted knowledge by measuring its impact on the performance of downstream tasks such as information retrieval or question answering. Finally, structure-based evaluation involves analyzing statistical properties of the extracted graph or structure—such as connectivity, depth, or presence of cycles—to assess its quality and consistency [40].

In recent years, with the rapid advancement of large language models (LLMs), their application in various NLP tasks such as text summarization, question answering, and sentiment analysis has been widely explored. Since these models have been trained on massive amounts of textual data, they have acquired knowledge about entities and the relationships between them. Given the challenges of extracted relations evaluation, particularly in the medical domain, where traditional evaluation methods are often difficult to apply, we employed an LLM-based approach to assess the correctness of extracted relations. Specifically, for each relation extracted from the text, we provided the LLM with the original sentence along with the two involved entities and asked it to determine whether a cause-effect or component-whole relation exists between them. If the LLM's assessment matched the existing relation in the relation set, the relation was retained; otherwise, if the LLM's judgment conflicted with the existing relation, the relation was removed from the set.

## IV. EMPIRICAL RESULTS

In this section, we review the quantitative and qualitative results obtained. We begin with the evaluation of the relation extraction model, followed by the assessment of the constructed relation set.

**System:**
You are an expert in biomedical language understanding. You carefully assess relationships between pharmaceutical entities based on the context of the sentence and your knowledge of drug interactions and classifications.

**user:**
Determine which of the two following semantic relationships best describes the relationship between the two highlighted drug entities in the sentence.

Available relation types:

1. cause-effect:
   - Definition: One drug causes, leads to, or affects the function or effectiveness of the other.
   - Example: "Combining warfarin with aspirin can increase the risk of bleeding."
     → warfarin <cause-effect> aspirin

2. component-whole:
   - Definition: One drug is a part or ingredient of a drug combination or formulation.
   - Example: "Vitamin B complex includes B1, B2, and B6."
     → B1 <component-whole> Vitamin B complex

Sentence: "{sentence}"
Entity 1 (e1): "{ent1}"
Entity 2 (e2): "{ent2}"

Choose the most suitable label: cause-effect or component-whole.
Only return one of these labels. No explanation needed.

Figure 4. Given prompt to model to validate extracted relations

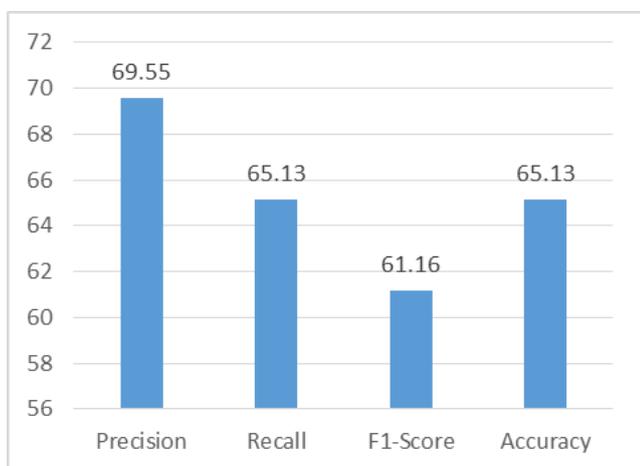

Figure 3. ACORD model evaluation on SemEval 2010 task8 dataset

## A. ACORD Relation Extraction Model Evaluation

In this study, the ACORD relation extraction model was trained on the SEMEVAL Task 8 2010 dataset, which includes 10 predefined relation labels: cause-effect, component-whole, content-container, entity-destination, entity-origin, instrument-agency, member-collection, message-topic, product-producer, and other. Given the importance of *cause-effect* and *component-whole* relations in the medical domain, we focused on these two as target classes and grouped the remaining eight relation types under the label *others*. Therefore, the ACORD model was trained as a three-class classifier aimed at identifying the relationship between two entities as either *cause-effect*, *component-whole*, or *others*. The model's performance based on various evaluation metrics is shown in Figure 3.

## B. Extracted Relations Evaluation

As previously mentioned, extracted relations were evaluated using a large language model. For this purpose, we employed the GPT-4o-mini model. The input prompt provided to the model is shown in Figure 4. The language model agreed with 71.48% of the extracted relations. Figure 5 presents the confusion matrix resulting from the comparison between the extracted relations and those identified by the LLM. As shown, extracted relations achieved a precision of 73%.

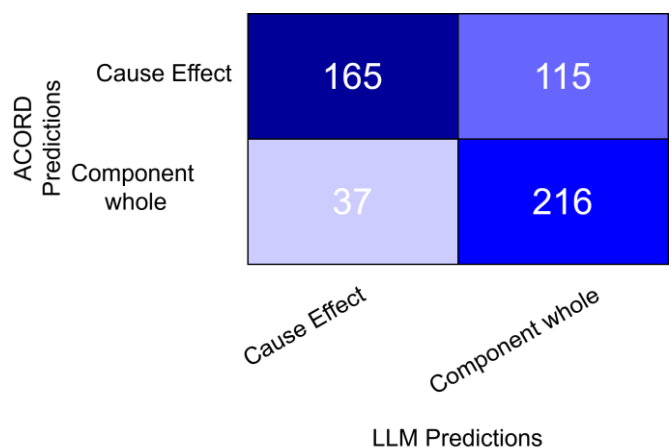

Figure 5. Confusion matrix of ACORD Predictions VS LLM Predictions

The set of relations before evaluation with the LLM consisted of 551 entities and 533 relations, of which 280 were related to cause-effect relations and 253 to component-whole relations. After evaluation with the LLM and removal of incorrect relations, the number of entities was reduced to 430 and the number of relations to 381. Figure 6 shows a portion of the set of relations in the form of an ontology. The full version of the relation set, including the complete ontology and all associated resources and code, is publicly available on GitHub[2].

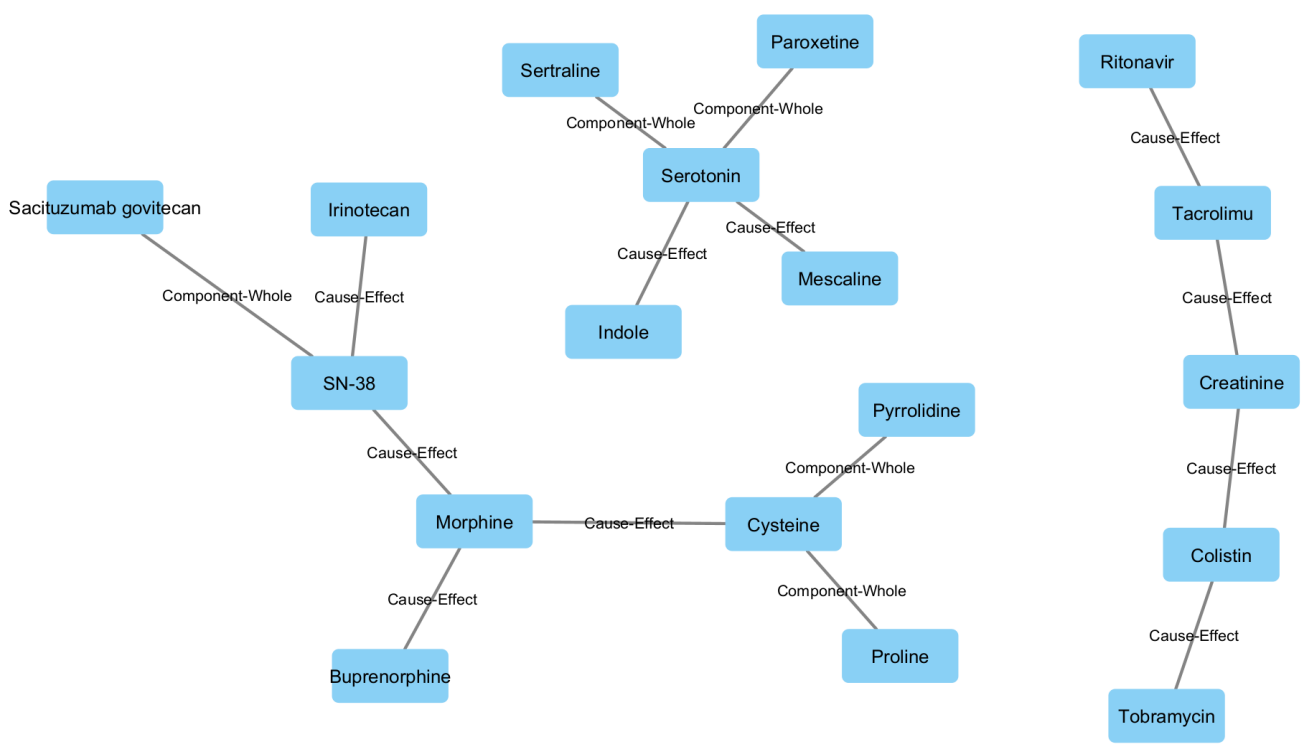

Figure 6. A representation of a portion of the extracted drug relations in the form of an ontology. Each node represents a drug, and each edge indicates one of the cause-effect or component-whole relations.

---

Table 1.Examination of ACORD–LLM conflict examples; "CE" denotes cause-effect and "CW" denotes component-whole.

| Sentence | ACORD | LLM |
|---|---|---|
| The presence of Cadmium, lead, nickel, Chromium, zinc, and mercury was also examined | CE | CW |
| The addition of both probiotic isolates, either broth or Wheat grains load separately has enhanced all the growth parameters; however, better results and increased production were in favor of adding probiotics with broth more than Wheat | CE | CW |
| Similarly, Histamine receptors in mast cells were significantly reduced after two different dosage of Bicalutamide treatment | CW | CE |
| The relationships among 5-HT3a receptor, Calcium/calmodulin (CaM) pathway, and ferroptosis were assessed via Western blotting, biochemical analysis, and lipid peroxidation assays, including iron and calcium content, reactive Oxygen species, glutathione peroxidase 4 (GPX4), ACSL, and CaM expression | CW | CE |

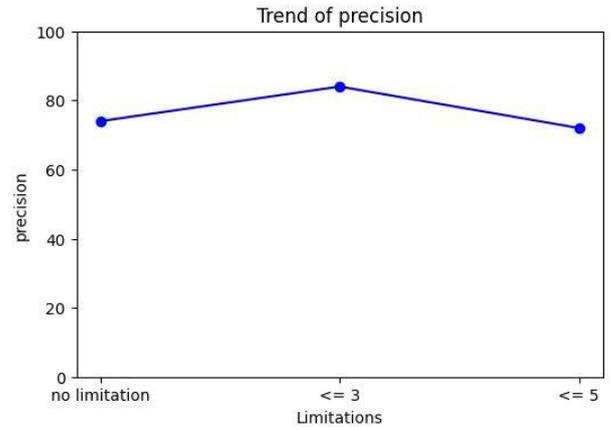

Figure 7. Changes in precision when restricting documents based on the number of words between two drugs. The best result was achieved when the shortest distance was considered.

## C. Interpretation of Non-Numerical Results

In this section, we provide a qualitative analysis of the results, focusing in particular on instances where the ACORD model's predictions conflicted with those of the LLM. Table 1 examines four such conflicts. One source of error is ambiguity: for example, in the first row of Table 1, the word "lead" appears as an element, but the relation extraction model treated it as a highly influential term in predicting a cause-effect relation. Additionally, since the relation extraction model identifies only drug entities, and we likewise limited our scope to drug entities, conflicts can arise when other non-drug entities intervene between two drugs and influence the true relationship—these cases were thus misclassified. Finally, the presence of identical drug entities appearing twice within the same text is another cause of disagreement between ACORD and the LLM; such examples are illustrated in subsequent rows of the Table 1.

As an additional analysis of our results, we hypothesized that a large distance between two entities might lead to reduced model accuracy. To evaluate this hypothesis, we conducted experiments under two different conditions. First, we selected sentences in which the distance between the two entities was at most three words, and then repeated the process for a maximum of five words. The precision obtained under three conditions—no restriction on word distance, a maximum distance of five words, and a maximum of three words—is shown in Figure 7. As illustrated, the best results were achieved when the entities were no more than three words apart. This suggests that, to ensure the reliability of extracted relations, it may be preferable to use only documents in which the entities are closely positioned. However, this approach cannot guarantee the completeness of the extracted relations.

## V. CONCLUSIONS AND FUTURE WORK

Relation extraction from text significantly aids in information analysis and classification. With the expansion of online repositories of medical articles, it has become possible to extract relationships between biomedical entities such as diseases. In this study, we present DREaM, which leverages the ACORD relation extraction model, applied to a subset of articles from the PubMed database, to identify drug–drug relations and create a set of such relationships. Since validating the correctness of extracted relations in the medical domain requires domain expertise, our approach employs large language models (LLMs) for this purpose. Relations identified by the ACORD model are included in the final set only if the LLM's assessment aligns with the model's prediction.

DREaM can be extended by incorporating additional entity types such as diseases and side effects. Moreover, its performance could be significantly improved by integrating a more advanced named entity recognition (NER) component. This framework can also be adapted to other domains—such as political, legal, and economic texts—to extract relations between domain-specific concepts.